\pdfoutput=1
\documentclass[11pt]{article}

\usepackage[]{acl} % review
\usepackage{subfigure}
\usepackage{xcolor,colortbl}
\usepackage{tabularx}
\usepackage{times}
\usepackage{latexsym}

\usepackage[T1]{fontenc}
\usepackage[utf8]{inputenc}

\usepackage{microtype}

\usepackage{amsmath,amssymb,amsfonts}
\usepackage{url,enumitem}
\usepackage{booktabs}
\usepackage{xspace}
\usepackage{graphicx}
\usepackage{comment}
\usepackage[normalem]{ulem}
\usepackage{mathtools}
\usepackage{svg}
\usepackage{xcolor, soul}% you can also write both the packages in this format  

\usepackage{algorithm,algorithmicx,algpseudocode}
\usepackage{listings}
\usepackage{makecell}

\newcommand{\methodname}[1]{\textsc{FluentPrompt}}
\newcommand{\newmethodname}[1]{\textsc{Unsupervised FluentPrompt}}
\newcommand{\autoprompt}[1]{$\text{AutoPrompt}_\text{SGD}$}

\usepackage{inconsolata}

\title{Toward Human Readable Prompt Tuning:\\ \textit{Kubrick's The Shining} is a good movie, and a good prompt too?}

\author{
Weijia Shi\thanks{\ \*  Equal contribution. Order randomly determined.} \qquad \qquad
        Xiaochuang Han\footnotemark[1]
        \\
        \textbf{Hila Gonen} \quad 
        \textbf{Ari Holtzman} \quad 
        \textbf{Yulia Tsvetkov} \quad
         \textbf{Luke Zettlemoyer}
 \\
   Paul G. Allen School of Computer Science \& Engineering, \\ University of Washington, Seattle, WA \\
  {\tt \{swj0419, xhan77, hilagnn, ahai, yuliats, lsz\}@cs.washington.edu}
}

\begin{document}
\maketitle
\begin{abstract}
% \weijia{Other title candidates: 
% 1. GPT learns from Kubrick's The Shining. Analyzing ... 
% 2. What makes Kubrick's the Shining a good prompt? Analyzing ....}
% \han{How about changing ReadPrompt to FluentPrompt}
% Language models have shown a remarkable ability to perform a new task by simply conditioning on a prompt. However, it remains unclear \textit{what} makes prompts work. In this work, we analyze \textit{which common attributes} of effective prompts share. Specifically, we propose a human readable prompt tuning method (\methodname{}) to find a diverse distribution of effective and fluent prompts that provides an important step for our analysis. 

%Language models have demonstrated their ability to perform downstream tasks with prompts. 
%With labeled training data, these prompts can usually be tuned by gradient-based methods. 
Large language models can perform new tasks in a zero-shot fashion, given natural language prompts that specify the desired behavior. Such prompts are typically hand engineered, but can also be learned with gradient-based methods from labeled data. However, it is underexplored \textit{what factors} make the prompts effective, especially when the prompts are natural language. In this paper, we investigate common attributes shared by effective prompts. We first propose a human readable prompt tuning method (\methodname{}) based on Langevin dynamics that incorporates a fluency constraint to find a diverse distribution of effective and fluent prompts. 
% It not only serves as an important step for our analysis, but also improves interpretability and transparancy of prompts. 
Our analysis reveals that \textbf{effective prompts are topically related to the task domain} and \textbf{calibrate the prior probability of label words}. Based on these findings, we also propose a method for generating prompts using only unlabeled data, outperforming strong baselines by an average of 7.0\% accuracy across three tasks. 
% Our experimental results show that this method can produce human-readable prompts that outperform the previous state-of-the-art zero-shot inference method by 5.7\% in accuracy. \weijia{maybe the abstract too short? what other details could we include?}

% Language models have shown a remarkable ability to perform a new task by simply conditioning on a prompt. However, it remains unclear \textit{what} makes prompts work. In this work, we analyze \textit{which common attributes} of effective prompts share. Specifically, we propose a human readable prompt tuning method (\methodname{}) to find a diverse distribution of effective and fluent prompts that provides an important step for our analysis. Our analysis shows that (1) \textbf{effective prompts are topically related to the task's domain} (2) \textbf{effective prompts calibrate the prior probability of the label words}. Inspired by our findings, we also propose a new method to generate effective prompts with only unlabelled data. Experimental results show that our proposed method can generate human readable prompts that outperforms previous zero-shot state-of-the-art inference method by 5.7\% in accuuracy. 

% \han{check title, whether need an intro figure, main claims, overall ready or not.}
% \han{highlight methodology as well (general), the analysis findings not so surprising}
% \han{grounding prior work analyzing prompt tuning}
% \han{make sure no overclaiming, reliable results}

% \han{are there any other human readable prompt tuning paper? check}

\end{abstract}

\section{Introduction}

Large language models can perform new tasks by simply conditioning on a prompt--a short sequence of text specific to the task. Such natural langauge prompts are either carefully hand engineered  (e.g., manual prompt engineering, \citealt{kojima2022large}) 
% \cite{kojima2022large, supernli, t0} 
or automatically learned from labeled data (e.g., gradient-based prompt tuning, \citealt{shin2020autoprompt}). 
% \cite{shin2020autoprompt, qin-eisner-2021-learning, openprompt, lester2021power}. 
Despite their effectiveness, it remains unclear what makes these prompts work and what attributes effective prompts share in common. 
In this paper, we aim to identify key characteristics of effective prompting, and use this knowledge to generate effective and human readable prompts without any labeled data.
% \han{use the findings to design an unsupervised objective for prompt tuning?}.

\begin{figure}[t]
\centering 
\includegraphics[scale=0.41]{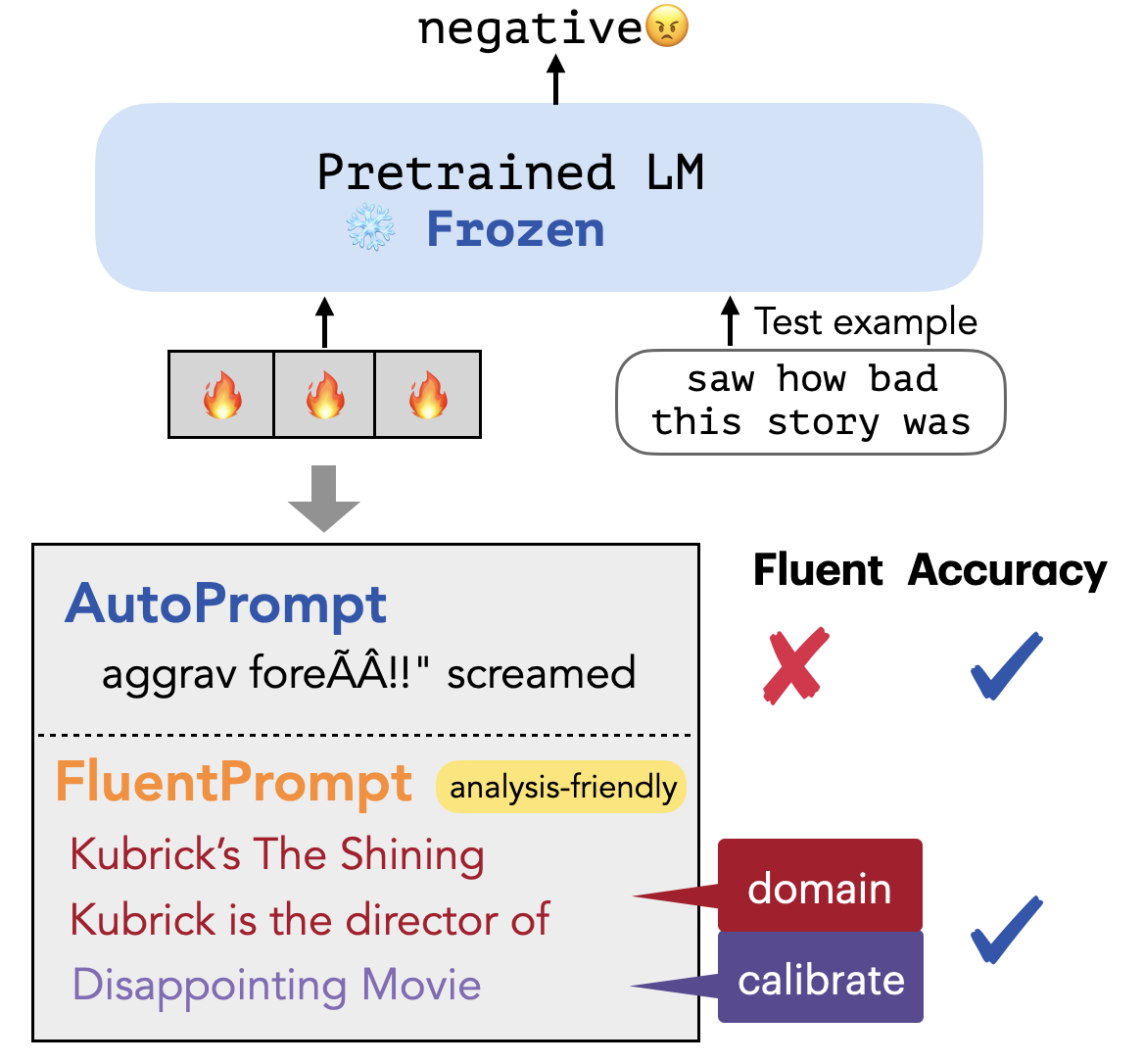}
\caption{
Compared with previous discrete prompt tuning method AutoPrompt \cite{shin2020autoprompt} which generates gibberish prompts, \methodname{} can identify effective and more readable prompts that are \textit{topically relevant to the task domain} and \textit{calibrate the prior probability of label words}.
} \label{fig:first}
\end{figure}

There are two main challenges for performing this type of analysis. First, manual prompt tuning produces a limited number of effective prompts for each task, making it difficult to infer common features of good prompts where contrast with less effective prompts is needed. 
% \weijia{not sure if the statement ma}
% in manually designed discrete prompts or discrete prompt tuning methods. This makes it difficult to perform correlational analysis. 
Additionally, the prompts found by gradient-based tuning methods are often disfluent and unnatural, making them difficult to interpret (e.g., AutoPrompt in \autoref{fig:first}). 

To overcome these challenges, we first propose a human readable prompt tuning method called \methodname{} based on a constrained decoding method. \methodname{} uses Langevin dynamics to generate a set of human readable prompts for any task. Our method adds a progressive noise to the tuning procedure to obtain a distribution of effective prompts, while also maintaining the fluency of the prompts through a perplexity constraint. 
% Our experimental results show that 
% As shown in Figure~\ref{fig:first}, compared with the baseline gibberish prompts generated by AutoPrompt, the prompts generated by \methodname{} are more readable (i.e., lower perplexity) and at the same time achieve competitive performance 
As shown in Figure~\ref{fig:first}, compared to the baseline gibberish prompts produced by AutoPrompt, \methodname{} generates prompts that are more fluent (i.e., lower perplexity) and perform competitively. 
% compared to prompts 
% generated by the baseline discrete prompt tuning method. 
The resulting fluent prompts not only facilitate our further analysis, but can also 
% improve the transparency of the model, leading 
lead to better trust and engagement from both researchers and end users. 
% an important advancement for bridging the gap between gradient-based prompt tuning and human written prompts.
% the use of readable prompts can improve the transparency and interpretability of the model, leading to better trust and engagement from both researchers and end users
% \han{say explicitly the method is targeting analysis}

After obtaining a diverse set of effective and human-readable prompts, we analyze the factors that contribute to the effectiveness of prompts. Specifically, we show that effective prompts are both (1) \textbf{topically related to the task domain} and (2) \textbf{more \emph{calibrated} to the task verbalizers.} Calibration measures how balanced the label word distribution of the prompted model is given an example-independent domain string~\cite{holtzman2019curious}. 
% Calibration measures how balanced the label word distribution of the prompted model is given an example-independent domain string. 
% We observse a strong correlation between the degree of ``calibration'' made by the prompt and model performance. 
% \weijia{give more clear definition of what is calibration}

Based on our findings, we propose a novel method \newmethodname{}, for automatically searching for effective prompts using only unlabeled data. \newmethodname{} optimizes the prompts for both better calibration and better domain relevance. Our experimental results show that \newmethodname{} outperforms 
% the previous state-of-the-art 
strong 
zero-shot baseline~\cite{holtzman-etal-2021-surface} by 7.0\% in accuracy. 
% , demonstrating its effectiveness
We summarize our contributions as follows:
% \hg{add references to sections?}
\begin{itemize}
    \item We introduce \methodname{}, a human-readable prompt tuning method that can generate a diverse set of \textit{effective} and \textit{fluent} prompts (\S \ref{sec:method}). This method not only serves as the foundation for our analysis, but also helps bridge the gap between manual prompt engineering and gradient-based prompt tuning. 
    \item We analyze the factors that contribute to the effectiveness of prompts and show that topic relatedness and calibration of the prompts are key to their success (\S \ref{sec:analysis}). 
    \item Inspired by our findings, we introduce a new method for discovering effective prompts without the need for labeled data (\S \ref{sec:new method}). 
\end{itemize}

\section{Related Work}
% \paragraph{Prompt Tuning}
% Prompt is a piece of text that is inserted before the input text, allowing language models to effectively perform a task. There are two main types of prompts: discrete and continuous. Discrete prompts are composed of discrete tokens from the vocabulary. Such prompt can be either human interpretable or uninterpretable. Previous works propose method to 

% Such prompt can be either discrete or continuous. 

% discrete prompt: human interpretable prompt found by prompt engineering and uninterpretable prompt found by autoprompt. 
% continuous prompt: continuous prompt tuning. not interpretable

% guides the language model to generate the target text. Such prompts can be descrete 

% Prompt tuning is a parameter-efficient method to adapt a pretrained language model to a downstream task 

% Prompt tuning refers to the process of adjusting the input prompts used in natural language generation systems in order to improve the quality and relevance of the generated output.

% The prompt is a short text that guides the language model to generate the target text. 

% As the scale of language model size increases, prompt tuning becomes more popular method to 

% discuss the figure and 

% \hg{maybe as a separate section?}
\subsection{Prompt Tuning}
% Prompt is a piece of text or continuous vector that is inserted before the input text, allowing language models to effectively perform a task.
% There are mainly two types of prompts: discrete and continuous prompt.

\paragraph{Continuous Prompt} 
Continuous prompts are continuous vectors
inserted to the task input for a prompted language model \cite{qin-eisner-2021-learning, openprompt, lester2021power,liu2021gpt}. Such continuous prompts are typically tuned by gradient-based methods, which are guided by the tasks training examples with labels. 
While these prompts usually improve the model performance, their continuous nature makes them difficult for humans to understand or interpret \cite{waywardness,hambardzumyan-etal-2021-warp}. 
% Different from discrete prompt, continuous prompts~ remove the retrictions that prompts are real token. Instead, continuous prompts, typically searched by gradient based method, can be mapped to any continuous vectors, which are difficult to understand and interpret~.

\paragraph{Discrete Prompt} 
Discrete prompts are composed of discrete tokens from natural language vocabulary. Such prompts can be either written by human or searched automatically. Human-written prompts \cite{kojima2022large, supernli, t0} typically consist of meaningful texts such as task descriptions \cite{schick-schutze-2021-exploiting} 
% \url{https://arxiv.org/pdf/2001.07676.pdf})
or instructions (e.g., ``let's think step by step'', \citealt{kojima2022large}), which are not only human readable but also aligns with their understanding of the task. In-context demonstration examples can also be considered as human-written prompts \cite{brown2020language, liu2022makes} but is not a focus of this work. 

Prior work has also focused on searching the discrete prompts automatically. One prominent way for this search can be gradient-based similar to the continuous prompt setup but with projections to a discrete vocabulary \cite{shin2020autoprompt}. The drawback of this method is that the resulting prompts are usually disfluent and difficult to read. 
% In contrast, previous work apply a gradient-based searched method~ over tokens in the vocabularty to automatically search for good prompts. Although the search is done in an automatic way, the resulting prompts are disfluent and hard to read (combination of random tokens). 
Other work searching for discrete prompts include edit-based enumeration \cite{prasad2022grips}, reinforcement learning \cite{deng2022rlprompt}, and large language model continuation and filtering \cite{zhou2022large}. 
The goal for these prompt tuning methods is mainly to achieve competitive task performance without modifying language model parameters. 

The purpose of our work is to analyze what aspects of the tuned natural language prompts make them effective for zero-shot language models. To facilitate such analysis, we need prompt readability as in human-written prompts and also a large search space as in gradient-based discrete prompt tuning. \methodname{} bridges the gap and provides a distribution of \textit{effective, diverse, and human-readable} prompts. 
% Our \methodname{} bridges the gap between human-written prompts and searched prompt by proposing a novel automatic method to search for human-readable prompts. Different from human-written or searched prompt that usually only provide a few effective prompt, \methodname{} provides a distribution of \textit{diverse, performant and human-readable} prompts which can facilitate further analysis.

\subsection{Analyses of Prompts}
A growing body of literature tries to understand the mechanisms behind prompts via various perspectives. 
For example, prompts in the form of in-context examples are analyzed under perturbations w.r.t. order, label, editing, etc. \cite{lu2022fantastically,min2022rethinking,chen2022relation}. 
Human-written instructions \cite{mishra2021cross} have also been studied and show weak sensitivity to semantic-changing perturbations \cite{webson2021prompt}. 
\citet{gonen2022demystifying} use paraphrasing and back-translation on a set of human-written prompts and analyze the correlation between their perplexity and performance. 
% shows that language models do not understand the meaning of prompts because ... We take a different approach showing that 

% Prior work also finds that model performance is highly sensitive to small changes in
% wordings (Mishra et al., 2022a) and that optimization over the discrete prompt space is non-trivial
% and often highly unstable.

% Our work not only propose a new human-readable prompt tuning method, but also provies 

% Do Prompt-Based Models Really Understand the Meaning of Their Prompts? study 
% Min shows that models learn just as well with incorrect labels as opposed to correct labels in in context learning

% Our work focuses on natural language prompts derived from gradient-based prompt tuning. \citet{waywardness} tune continuous prompts and investigate their projections to the discrete vocabulary space. In contrast, we perform prompt tuning in the discrete space directly with \methodname{}, for a faithful interpretation of the prompt that remains the same for inference and analysis. 

Our work focuses on natural language prompts derived from gradient-based prompt tuning. \citet{waywardness} tune continuous prompts and shows effective continuous prompts may transfer poorly to their nearest discrete prompts. 
% , suggesting that differentiable interpretable-prompt
% optimization remains a significant challenge.  
In contrast, we perform prompt tuning in the discrete space directly with \methodname{}, demonstrating the feasibility of searching for readable prompts using gradient-based method. 
% shed lights on a deeper understanding of what makes natural language prompts work.
This approach gives us a more faithful understanding of 
 % Through this approach, we aim to gain a deeper understanding of 
 the factors that contribute to the effectiveness of natural language prompts.

\section{\methodname{}} \label{sec:method}
% \section{Finding Prompts / Obtaining prompts for analysis} 
% \label{sec:method}
% We propose a novel method -- \methodname{} -- aiming to find a distribution of human-readable prompts. 
% From a quantitative perspective, a distribution of prompts ranging in different performance is desired rather than a single best-performing prompt, to discover potential correlation between the features of the prompts and their performance. 
% From a qualitative perspective, readable prompts can facilitate better understanding and trust from human researchers and users. 

\methodname{} generates a diverse set of human-readable prompts. Our goal is not only to identify a single best-performing prompt, but also to explore the relationship between the features of the prompts and their performance.

% We believe that this approach has two main benefits. First, from a quantitative perspective, examining a range of prompts with varying levels of performance allows us to gain a deeper understanding of how different prompts behave in different contexts. Second, from a qualitative perspective, the use of readable prompts can improve the transparency and interpretability of the model, leading to better trust and engagement from both researchers and end users. 

% To find a distribution of prompts with a range of performance is for quantitative analysis, and to find human readable prompts is for qualitative understanding. 
\subsection{Background: continuous prompt tuning}
% \hg{I think this part would be clearer if you explain in 1-2 sentences the high level idea, before the actual details} 
%Language models have been \emph{prompted} to perform downstream tasks without finetuning model parameters (e.g., \citealt{schick-schutze-2021-exploiting}). 
%In this work, we consider an autoregressive language model $\theta$ that can be prompted to perform a text classification task. 
Given an input example $\boldsymbol{x}$ with an output label $y \in Y$, we can {\em prompt} an autoregressive language model with parameters $\theta$ as follows. We reformulate the task as a language modeling problem by inserting a task-specific template $\boldsymbol{t}$ to $\boldsymbol{x}$ and defining a verbalizer $v$ mapping from a label $y$ to a label word (i.e, token in the LM's 
% \hg{do we say earlier that a language model is LM? should be in paranthesis in the first time a language model is mentioned} 
vocabulary). The probability of the label is estimated by: 
% \begin{align*}
% % \small
% p_{\theta}(y \mid \boldsymbol{x}) \propto  \text{p}_{\theta} (v(y) 
% \mid \boldsymbol{x}, t)   
% % \frac{EXP \text{p}_{\theta}(x_{\text{next}}}
% \end{align*}
\begin{align*}
p_{\theta} (v(y) 
\mid \boldsymbol{x}, \boldsymbol{t}) = 
\frac{\exp \text{logit}_{\theta}(v(y) \mid \boldsymbol{x}, \boldsymbol{t})}{\sum_{y'} \exp \text{logit}_{\theta}(v(y') \mid \boldsymbol{x}, \boldsymbol{t})}
\end{align*} 
\citet{lester2021power} add a sequence of $M$ soft embeddings 
$\Tilde{\boldsymbol{e}}_{0:M}$ (simplified as $\Tilde{\boldsymbol{e}}$; 0:$M$ refers to the positional subscript for the sequence from position 0 to $M-1$)
% $\Tilde{\boldsymbol{e}}$
in front of the input. Therefore, the probability of the label is computed by $p_{\theta}(v(y) \mid \Tilde{\boldsymbol{e}}, \boldsymbol{x}, \boldsymbol{t})$, where $\Tilde{\boldsymbol{e}}$ is embeddings that bypass the word embedding layer of the LM $\theta$ and is learned based on a set of training data. These learned embeddings are sometimes referred to as soft prompts, and the learning of such prompts as soft prompt tuning. For example, if stochastic gradient descent (SGD) is used as an optimizer, the soft prompt $\Tilde{\boldsymbol{e}}$ is updated as
\begin{align*}
    \Tilde{\boldsymbol{e}}^{i} = \Tilde{\boldsymbol{e}}^{i-1} - \eta \nabla_{\Tilde{\boldsymbol{e}}} (- \log p_{\theta}(v(y) \mid \Tilde{\boldsymbol{e}}^{i-1}, \boldsymbol{x}, \boldsymbol{t}))
\end{align*}
where $i$ is the timestep superscript, referring to $i$-th optimization step. 

\subsection{Discrete prompt tuning with Langevin dynamics} \label{sec: langevin}
% There are two problems with the above soft prompt tuning: (1) the resulting embeddings do not represent the natural language vocabulary, and (2) it gives us one embeddings instead of a distribution of such embeddings with a range of performance. This hinders a potential analysis of why certain prompts are helpful to the LM in given tasks, specifically in (1) discovering features of the prompts and (2) understanding their comparative effect (performance).
There are two challenges for the soft prompt tuning reviewed above. First, the resulting embeddings cannot be mapped to the natural language vocabulary. \citet{waywardness} show that naively mapping an effecive soft prompt to their nearest tokens significantly drops the performance.
Second, we only obtain a single embedding instead of a range of embeddings with varying levels of performance. This makes it difficult to analyze the characteristics of the prompts and compare their effectiveness in specific tasks for the language model.

% The soft prompt tuning method has two problems. Firstly, the resulting embeddings cannot be easily matched to the natural language vocabulary. Research has shown that simply mapping an effective soft prompt to the closest tokens significantly decreases performance. Secondly, we only get a single embedding instead of a distribution of embeddings with a range of performance levels. This impedes our ability to analyze the features of the prompts and understand their comparative effect on the language model's performance in specific tasks.
% Both hindering an extensive analysis of why certain prompts are helpful to the LM in given tasks. 

% \hg{I think it should be clearer in this part - what is your contribution vs. what has been done before 
% (this work is not the first to do discrete prompt tuning, and this is hard to tell, especially given that this title doesn't contain "background", unlike the previous section) - it is mentioned in the last paragraph but I guess it should be done earlier? Why not describe autoprompt as background and then state what changes are made on top of it?}

Following \citet{kumar2022gradient}, we use Langevin dynamics to sample discrete prompts that lead to a better performing model in the task. 
% in a discrete representation space
Overall, the method is similar to SGD but adds a progressive Gaussian noise to the embeddings, with the scale decreasing over time. Additionally, at each optimization step, the updated embedding is projected to the nearest embedding in the LM vocabulary. 
%This ensures each token in the tuned prompt corresponds directly to natural language vocabulary. 
\begin{align*}
    \Tilde{\boldsymbol{e}}^{i} = \text{Proj}_{\mathbf{E}}[ &\Tilde{\boldsymbol{e}}^{i-1} - \eta \nabla_{\Tilde{\boldsymbol{e}}} \mathcal{E}(\Tilde{\boldsymbol{e}}^{i-1}) + \sqrt{2 \eta \beta_i} \boldsymbol{z}]
\end{align*}
where:
\begin{itemize}
    \item $\mathcal{E}$ is an energy function (lower is better), $\mathcal{E}(\Tilde{\boldsymbol{e}}^{i-1}) = - \log p_{\theta}(v(y) \mid \Tilde{\boldsymbol{e}}^{i-1}, \boldsymbol{x}, \boldsymbol{t})$.
    \item $\boldsymbol{z}$ is a Gaussian noise, $\boldsymbol{z} \sim \mathcal{N}(0, I_{|\Tilde{e}|})$.
    \item $\beta$ is the variance of the noise following a geometric progression, $\beta_{\text{start}} > \beta_i > \beta_{\text{end}} \xrightarrow{} 0$.
    \item $\mathbf{E}$ is the embedding table (layer) of the LM $\theta$, one embedding for each token in the vocabulary.
    \item $\text{Proj}_\mathbf{E}$ is a projection operation finding a nearest neighbor for each soft embedding in the LM's vocabulary, $\text{Proj}_\mathbf{E}(\Tilde{e}) = \text{argmin}_{e_v \in \mathbf{E}}(\lVert e_v - \Tilde{e} \rVert_2)$.
\end{itemize}

Without the progressive noise in Langevin dynamics, our prompt search procedure is gradient-based and shares a similar intuition with AutoPrompt \citep{shin2020autoprompt}. Both methods use the gradient of the loss w.r.t. the embeddings, though AutoPrompt applies greedy substitution whereas we use projected gradient descent, aligning with soft prompt tuning and enabling the subsequent prompt sampling. AutoPrompt also incorporates verbalizer word selection, which is not a focus of the analysis in this work. We use our gradient-based, discrete prompt tuning method without Langevin dynamics as a baseline, referred to as \autoprompt{}. 

\subsection{Fluency constraint}
\label{sec:fluency_constraint}
Sampling from projected Langevin dynamics ensures that the tuned prompt contains natural langauge tokens. However, with no extra constraints, they can form a disfluent sentence. 

We explicitly incorporate a fluency objective to the Langevin energy function. This objective resembles the regular perplexity loss, but the labels (next token in the prompt) are not ground-truth. Instead, we measure an embedding-based sequence probability according to \citet{kumar2022gradient}. For simplicity, below we drop the timestep superscript on the prompt embeddings and only keep the positional subscript. 

% The first step is to obtain the probability of generating the embedding at position $m$ with the previous $m-1$ embeddings. We extract the output embedding from the LM (before converted to output logits) at position $m-1$: $h_{\theta, m-1} = h_{\theta}(\Tilde{\boldsymbol{e}}_{0:m})$. Then the probability is: 
The first step is to obtain the probability of generating the embedding at position $m$ (i.e., $\Tilde{e}_{m}$) based on the previous $m-1$ embeddings (i.e., $\Tilde{\boldsymbol{e}}_{0:m}$). We extract the last hidden state from the LM (i.e., output embedding) at position $m-1$: $h_{\theta, m-1}=h_{\theta}(\Tilde{\boldsymbol{e}}_{0:m})$.
% = h_{\theta}(\Tilde{\boldsymbol{e}}_{0:m})$. 
Then the probability is: 

% \han{we are dropping the iteration/timestep indicator $i$ for simplicity below; also mention that the perplexity defined below for the embeddings can take the input x or template t as conditions (in the case of tail prompt)}
\begin{align*}
    p_{\theta}(\Tilde{e}_{m} \mid \Tilde{\boldsymbol{e}}_{0:m}) = \frac{\exp (h_{\theta, m-1} \cdot \Tilde{e}_{m})}{\sum_{e_v \in \mathbf{E}} \exp (h_{\theta, m-1} \cdot e_v)}
\end{align*}
where we equivalently compute the logits for each embedding's corresponded vocabulary and take the softmax.\footnote{This is equivalently computing the logits since $e_v$ and the projected $\Tilde{e}_{m}$ from the last optimization step are both in the embedding table.} 
Subsequently, the sequence probability is $p_{\theta}(\Tilde{\boldsymbol{e}}_{0:M}) = \prod_{m=1}^{M-1} p_{\theta}(\Tilde{e}_{m} \mid \Tilde{\boldsymbol{e}}_{0:m})$. 
% \han{check in our implementation whether there's a learned bias term}

We define a prompt fluency loss as the negative log-likelihood of the prompt embeddings, $-\log p_{\theta}(\Tilde{\boldsymbol{e}}_{0:M})$. Along with the task labeling loss (\S \ref{sec: langevin}), we modify our energy function as:
\begin{align*}
    \mathcal{E}(\Tilde{\boldsymbol{e}}_{0:M}) = &- \lambda_{\text{task}} \log p_{\theta}(v(y) \mid \Tilde{\boldsymbol{e}}_{0:M}, \boldsymbol{x}, \boldsymbol{t})\\
    &- \lambda_{\text{fluency}} \log p_{\theta}(\Tilde{\boldsymbol{e}}_{0:M})
\end{align*}
where $\lambda_{\text{task}} + \lambda_{\text{fluency}} = 1$. Through the whole \methodname{} tuning procedure, the language model parameters $\theta$ is fixed while the embeddings $\Tilde{\boldsymbol{e}}_{0:M}$ are tuned.

\subsection{Experimental Setup}
% \section{Experiments}
% \subsection{Experimental Setup}

\paragraph{Target tasks} 
We evaluate performance on two sentiment analysis tasks: Amazon Polarity~\cite{mcauley2013hidden} and SST-2~\cite{socher-etal-2013-recursive}, and one topic classification task: AGNEWS~\cite{Zhang2015CharacterlevelCN}. 
These tasks were selected since vanilla soft prompt tuning~\cite{lester2021power} substantially improves model performance. 
In contrast, tasks like RTE~\cite{dagan2005pascal} are more difficult; soft prompt tuning did not yield a significant improvement (57.4\% accuracy from prompt tuning compared with 52.1\% from random guess) in our pilot study, and we therefore did not pursue further analysis using \methodname{}. 
% \han{added lines above, does it make sense? probably move to footnote?} 
The verbalizer words and templates used for each task are listed in \autoref{tbl:verbalizer}. 
\begin{table}[t]
\centering
\small
\begin{tabular}{lcl}
\toprule
% \midrule
\textit{Prompt} &\textit{Acc.} & \textit{PPL} \\
\midrule
\multicolumn{3}{c}{\colorbox{blue!20}{SST-2}} \\
\textbf{Empty Prompt} & 66.5 & - \\
\textbf{\autoprompt} &  &  \\

\hspace*{1mm} Compl disgustingÃÂÃÂ Rated jer & 87.6 & $>10^6$  \\
% \midrule
\textbf{\methodname} & & \\
% \midrule
% \sethlcolor{pink!10}  
\hspace*{1mm} \textcolor{red}{Kubrick, "The Shining} & 87.5 & 13.1 \\
% \hspace*{1mm} Paramount, "The Shining & 86.8 & 12.2 \\
% \hspace*{1mm} Kubrick\textbackslash 's "The Man & 86.3 & 9.3 \\
% \hspace*{1mm} Kubrick\textbackslash 's "The Last & 84.8 & 10.7 \\
\hspace*{1mm} \textcolor{red}{Paramount, "The Shining} & 86.8 & 12.2 \\
\hspace*{1mm}  \textcolor{red}{Kubrick\textbackslash 's "The Man} & 86.3 & 9.3 \\
\hspace*{1mm} \textcolor{blue}{disappointing.\textbackslash n\textbackslash n"} & 84.4 & 4.1 \\
% \hspace*{1mm} \textcolor{blue}{complained.\textbackslash n\textbackslash n"} & 83.6 & 3.4 \\

\midrule
\multicolumn{3}{c}{\colorbox{blue!20}{AMAZON}} \\
% \midrule
\textbf{Empty Prompt} & 75.8 & - \\
\textbf{\autoprompt} &  &  \\
\hspace*{1mm} Reviewed experien audition lashesrible & 82.2 & $>10^6$ \\
\textbf{\methodname} & & \\
\hspace*{1mm} \textcolor{blue}{scathing.\textbackslash n\textbackslash n"} & 83.1 & 5.1 \\
 % \hspace*{1mm} Yelp Yelp. I ordered & 82.7 & 46.4 \\
% \hspace*{1mm} this is disgusting." & 82.6 & 4.2 \\
\hspace*{1mm} \textcolor{blue}{upset.\textbackslash n\textbackslash n"} & 82.6 & 3.67 \\
% \hspace*{1mm} Weinstein.\textbackslash n\textbackslash n" & 82.5 & 4.85 \\
\hspace*{1mm} \textcolor{red}{cigars: \textbackslash n\textbackslash n} & 82.4 & 20.9 \\
\hspace*{1mm} \textcolor{red}{mascara\textbackslash n\textbackslash n\textbackslash n} & 82.2 & 47.1 \\
% \hspace*{1mm} ugly.\textbackslash n\textbackslash n" & 82.2 & 4.7 \\
% \hspace*{1mm} mascara is\textbackslash n\textbackslash n & 81.5 & 51.6 \\

\midrule
\multicolumn{3}{c}{\colorbox{blue!20}{AGNEWS}} \\
% \midrule
\textbf{Empty Prompt} & 49.7 & - \\
\textbf{\autoprompt} &  &  \\
\hspace*{1mm} EStreamFramenetflixnetflixobookgenre & 69.3 & $>10^5$ \\

\textbf{\methodname} & & \\
% \hspace*{1mm} netflix.com/browse/genre/4 & 69 & 1.69 \\
\hspace*{1mm}  \textcolor{red}{netflix/genre/netflix} & 71.1 & 281.0 \\
\hspace*{1mm} \textcolor{red}{netflix AnimeMoviegenre\textbackslash n} & 70.1 & 1925.0 \\
\hspace*{1mm}  \textcolor{red}{Synopsis\textbackslash n\textbackslash nThe story is} & 69.2 & 9.6 \\

% \hspace*{1mm} netflix.com/WiMovie/7017867 & 68 & 3.66 \\
% \hspace*{1mm} netflix.com/netflix/ & 67 & 3.15 \\

\hspace*{1mm}  \textcolor{red}{pmwiki.php/main/Superhero} & 65.0 & 2.4 \\
% \hspace*{1mm}  \textcolor{red}{genre/pmwiki/pmwiki} & 64.4 & 57.2 \\

% \hspace*{1mm} pmwiki/pmwiki.php/Manga/K & 64.5 & 2.75 \\
% \hspace*{1mm} netflix/pmwiki/pmwiki & 64.1 & 3.7 \\

\bottomrule

\end{tabular}
\caption{Accuracy (\textit{Acc.}) and Perplexity (\textit{PPL}) of prompts. Both \methodname{} and \autoprompt{} use $M$=5 tunable tokens. \methodname\~ shows comparable performance to the \autoprompt{} but with significantly lower perplexity. Prompts discovered by \methodname{} show \textcolor{red}{domain relevance} and potential \textcolor{blue}{caliberation} for model outputs. 
% consist of \textcolor{red}{domain words} and \textcolor{blue}{words used for calibration}. 
} \label{tbl:prompt}

\end{table}

\begin{table*}[]
\small
\centering
\begin{tabular}{@{}lccccccccc@{}}
\toprule
{} & \multicolumn{2}{c}{\textbf{SST-2}} & \multicolumn{2}{c}{\textbf{Amazon}}& \multicolumn{2}{c}{\textbf{AGNews}} \\ \cmidrule(l){2-3} 
\cmidrule(l){4-5} 
\cmidrule(l){6-7} 
& \multicolumn{1}{c}{\begin{tabular}[c]{@{}c@{}}$\log(\text{perplexity})$\end{tabular}} & \multicolumn{1}{c}{\begin{tabular}[c]{@{}c@{}}Accuracy\end{tabular}} &   \multicolumn{1}{c}{$\log(\text{perplexity})$} & \multicolumn{1}{c}{\begin{tabular}[c]{@{}c@{}}Accuracy\end{tabular}} &   \multicolumn{1}{c}{$\log(\text{perplexity})$} & \multicolumn{1}{c}{Accuracy}   \\ \midrule
$\lambda_{\text{fluency}}=0$ & 13.75 {\tiny $\pm$ 1.81} & 87.55 {\tiny $\pm$ 0.95}   & 14.32 {\tiny $\pm$ 1.31}  & 75.31 {\tiny $\pm$ 1.76}    & 15.04 {\tiny $\pm$ 3.30} & 74.56 {\tiny $\pm$ 1.65} \\ 
$\lambda_{\text{fluency}}=0.003$ & 9.86 {\tiny $\pm$ 2.41} & 88.86 {\tiny $\pm$ 0.67}   & 10.44 {\tiny $\pm$ 2.09}  & 86.37 {\tiny $\pm$ 0.68}    & 10.13 {\tiny $\pm$ 1.13} & 76.43 {\tiny $\pm$ 1.05}  \\ 
\bottomrule
\end{tabular}
\caption{Accuracy and perplexity of the prompts tuned with and without the readability constraint $\lambda_{\text{fluency}}$. For $\lambda_{\text{fluency}} > 0$, we report the best value ($\lambda_{\text{fluency}}=$0.003) across 4 learning rates and 5 random seeds with $M=$10. All $t$-tests of perplexity and accuracy show $p \leq$ 0.0001. 
% Han: checked, 20 prompts used in every scenario, likely across 4 lr, 5 seeds
}
\label{tab:effect_of_lambda}
\end{table*}

\paragraph{Model} We optimize prompts for GPT-2 large (774M parameters, \citealt{radford2019language}) using \methodname{}. We use a batch size of 16 and train for 5,000 steps with an AdamW optimizer \citep{loshchilov2018decoupled}. We select the best prompt based on the validation performance. 
For our method \methodname{}, we use a learning rate $\eta \in$ \{0.3, 1.0, 3.0, 10.0\}, $\beta_{\text{start}} =$ 1.0, $\beta_{\text{end}} =$ 0.0001, $\lambda_{\text{fluency}} \in$ \{0.003, 0.01, 0.03, 0.1, 0.3\}. We search for both 5-token prompts ($M=5$) and 10 token ($M=10$) prompts and use five random seeds for each hyperparameter setup. 
% \hg{how do you use the 5 different result sets?} 
Additionally, we perform experiments with $\beta_{\text{start}} = \beta_{\text{end}} = 0$ (i.e, no progressive noise) and $\lambda_{\text{fluency}} = 0$ (i.e, no fluency constraint) as ablations 
% \hg{ablation models/experiments?} 
to \methodname{} purposed for analysis. 
% \han{check whether we used AdamW optimizer}
% \weijia{maybe mention this in 4? Should also add autoprompt}

% \paragraph{Baseline}

% \paragraph{Baselines}
% \han{Describe Autoprompt and our pure gradient method}

% \subsection{Results}
% \weijia{which one do we use?}

% \subsection{Advantages of \methodname{}} 
\subsection{Results} 
% \hg{the title is a bit weird, I think. maybe something like "advantages of readprompt", or something about the components of readprompt?}
% \weijia{discuss results}
% \han{rename AutoPrompt}

Table~\ref{tbl:prompt} shows the accuracy and perplexity of empty prompt (i.e., no $\Tilde{\boldsymbol{e}}$), \autoprompt{} and \methodname{}, along with example prompts for each method. 
We see that \methodname{} performs comparably to \autoprompt{} and significantly better than the empty prompt. In terms of readability, \methodname{} generates more fluent prompts than \autoprompt{}. 
% This suggests that \methodname{} is a significant step towards finding fluent natural language prompts that also have strong performance, which 
%This is an important advancement for our further analysis. 
% \hg{maybe you should state that it is an important advancement not only for analysis? this feels pretty modest.}

To further understand the contributions of \methodname{}, we now investigate the effects of its two key modifications on top of \autoprompt{}: the noise $\boldsymbol{z}$ in Langevin dynamics and the weight $\lambda_{\text{fluency}}$ for prompt fluency.
% \methodname{} proposes two key modifications to the regular gradient-based prompt tuning\weijia{unify the name: \autorpompt}, the noise $\boldsymbol{z}$ in Langevin Dynamics and a weight $\lambda_{\text{fluency}}$ for prompt fluency. 
% In this section, we investigate the effect of each modification. 

% \han{TODO for the tables: check postprocessing again to match the amount of used lambda, learning rate, random seeds, etc. for both setups}
% \han{Mention the experiments in the section iterate through different $z$ and $\lambda$, and we investigate the effect of $z$ and $\lambda$ separately.}

\paragraph{Effect of $\lambda_{\text{fluency}}$}
% \hg{the first part of the sentence hints that the assumption should be that this constraint would hurt performance. why would we assume that? I'd remove the first part and just state that it improves}
% Instead of compromising prompt performance by adding a constraint, the readability constraint weighted by $\lambda_{\text{fluency}}$ improves performance and helps in finding lower-perplexity prompts. 
In \autoref{tab:effect_of_lambda} we show the performance with and without the fluency constraint ($\lambda_{\text{fluency}} = 0.003$ and $\lambda_{\text{fluency}} = 0$) and the log-perplexity of the discovered prompts. The fluency constraint effectively leads to significantly lower perplexity and also better accuracy ($p \leq$ 0.0001 in all $t$-tests).\footnote{On human-written prompts, \citet{gonen2022demystifying} report a similar finding.} Prompts with lower perplexity are desired for their potentially better readability for downstream analyses. 
% \hg{you can actually cite my paper here, which supports this argument directly, I've recently uploaded to arxiv: https://arxiv.org/pdf/2212.04037.pdf. not a must of course :)}
% \han{compared $\lambda$=0 and $\lambda$=0.003 by looking at a t-test on group acc and ppl.}

% $30.0 \pm 1.0_{(p=0.02)}$

\paragraph{Effect of $\boldsymbol{z}$}
The progressive noise $\boldsymbol{z}$ helps find a diverse set of prompts while not compromising overall performance. 
In \autoref{tab:effect_of_z} we show the best and average prompt performance with and without the noise $\boldsymbol{z}$ (i.e., $\beta > 0$ and $\beta = 0$). We measure the diversity of prompts by Dist-1, a unigram distinctiveness metric \citep{li-etal-2016-diversity}. We find that the prompts obtained with $\boldsymbol{z}$ ($\beta > 0$) are more diverse and overall have an on-par performance with the setup without $\boldsymbol{z}$ ($\beta = 0$).

\begin{table*}[t]
\small
\centering
\begin{tabular}{@{}lccccccccc@{}}
\toprule
{} & \multicolumn{3}{c}{\textbf{SST-2}} & \multicolumn{3}{c}{\textbf{Amazon}}& \multicolumn{3}{c}{\textbf{AGNews}} \\ \cmidrule(l){2-4} 
\cmidrule(l){5-7} 
\cmidrule(l){8-10} 
& \multicolumn{1}{c}{\begin{tabular}[c]{@{}c@{}}Max\end{tabular}} & \multicolumn{1}{c}{\begin{tabular}[c]{@{}c@{}}Mean\end{tabular}} & \multicolumn{1}{c}{Dist-1} & \multicolumn{1}{c}{Max} & \multicolumn{1}{c}{\begin{tabular}[c]{@{}c@{}}Mean\end{tabular}} & \multicolumn{1}{c}{Dist-1} & \multicolumn{1}{c}{Max} & \multicolumn{1}{c}{Mean} & \multicolumn{1}{c}{Dist-1} \\ \midrule
$\beta=0$ & 90.2 & 86.5  & 72.6 & 87.7  & 85.1   & 57.9 & 82.6 & 71.6 & 81.7 \\ 
$\beta>0$ & 89.6 & 85.5  & 77.6 & 88.7  & 85.4   & 61.2 & 80.7 & 74.1 & 82.2 \\ 
% (Bartlett) &  &   & B=47.7 &  &   & B=34.7 &  &  & B=59.3 \\ 
%  &  &   & p<0.0001 &  &   & p<0.0001 &  &  & p<0.0001 \\ 
\bottomrule
\end{tabular}
\caption{Prompt performance and diversity with and without the progressive noise $\boldsymbol{z}$ ($\beta > 0$ and $\beta = 0$).}
\label{tab:effect_of_z}
\end{table*}

\section{What makes good prompts?} \label{sec:analysis}
% In this section, we examine what attributes of prompts lead to good performance and transferrability by analyzing the set of prompts \hg{we obtain using \methodname{}}. %we found in Section~\ref{sec:method}.
% We use SST-2 and ... for visualization...
% In this section, we examine what attributes of prompts lead to good performance. Specifically, we study the 10-token prompts found by \methodname{} on SST-2, Amazon and AGNEWS. 
In this section, we analyze common attributes of the effective tuned prompts. Specifically, we study the 10-token prompts found by \methodname{} on SST-2, Amazon and AGNEWS. 

\begin{figure*}
    \centering
    \subfigure[]{\includegraphics[width=0.325\textwidth]{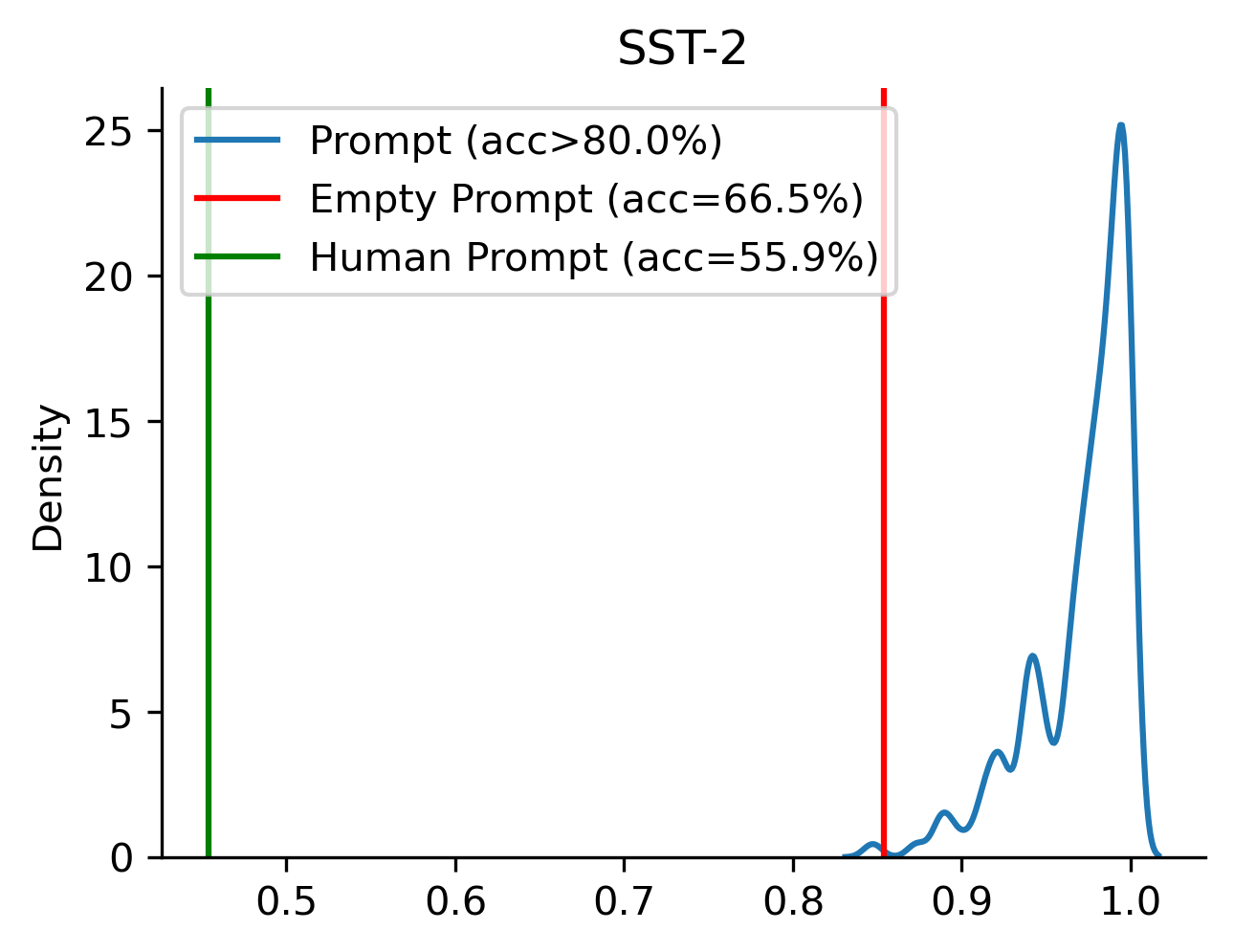}} 
    \subfigure[]{\includegraphics[width=0.325\textwidth]{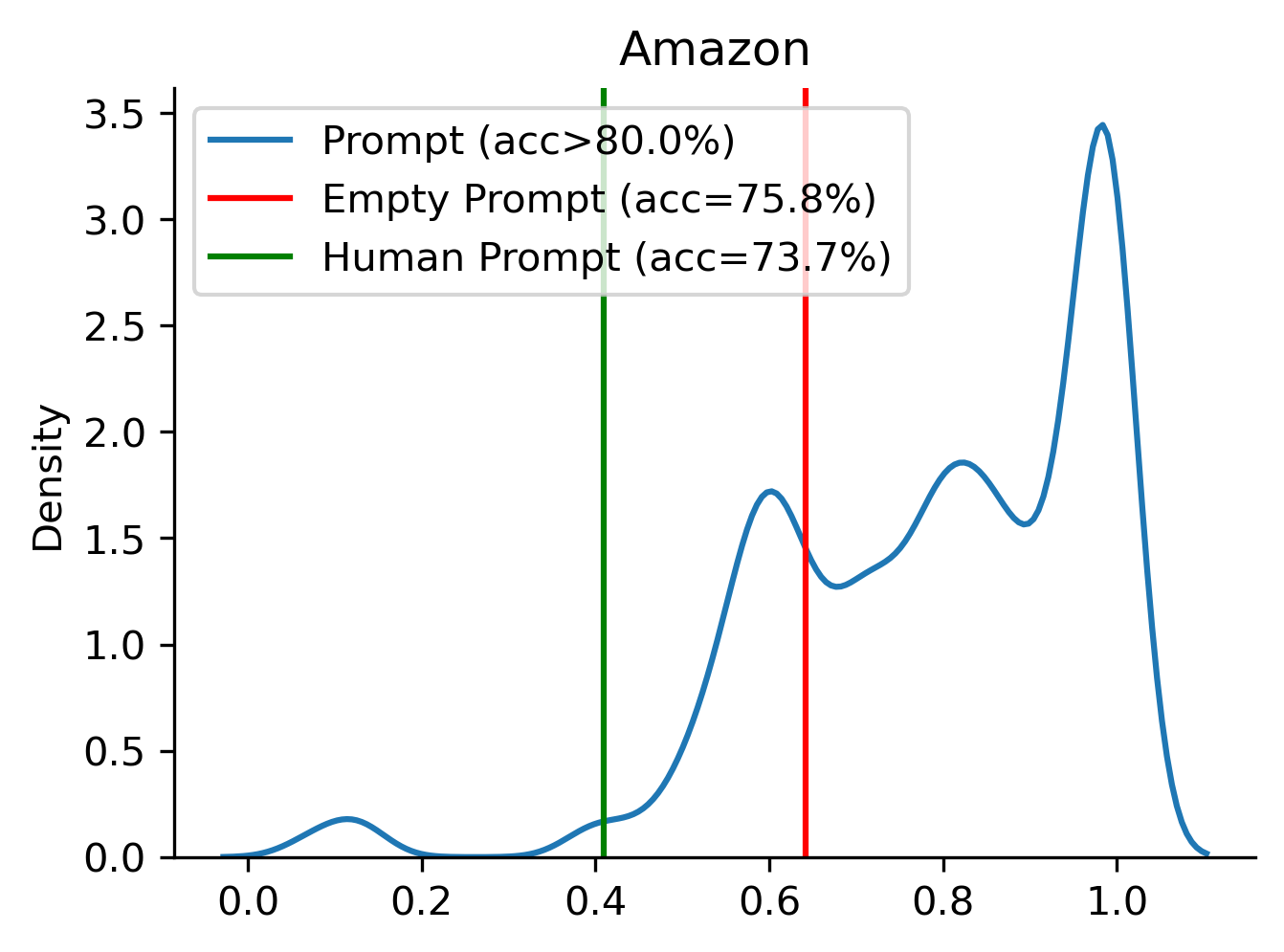}} 
    \subfigure[]{\includegraphics[width=0.325\textwidth]{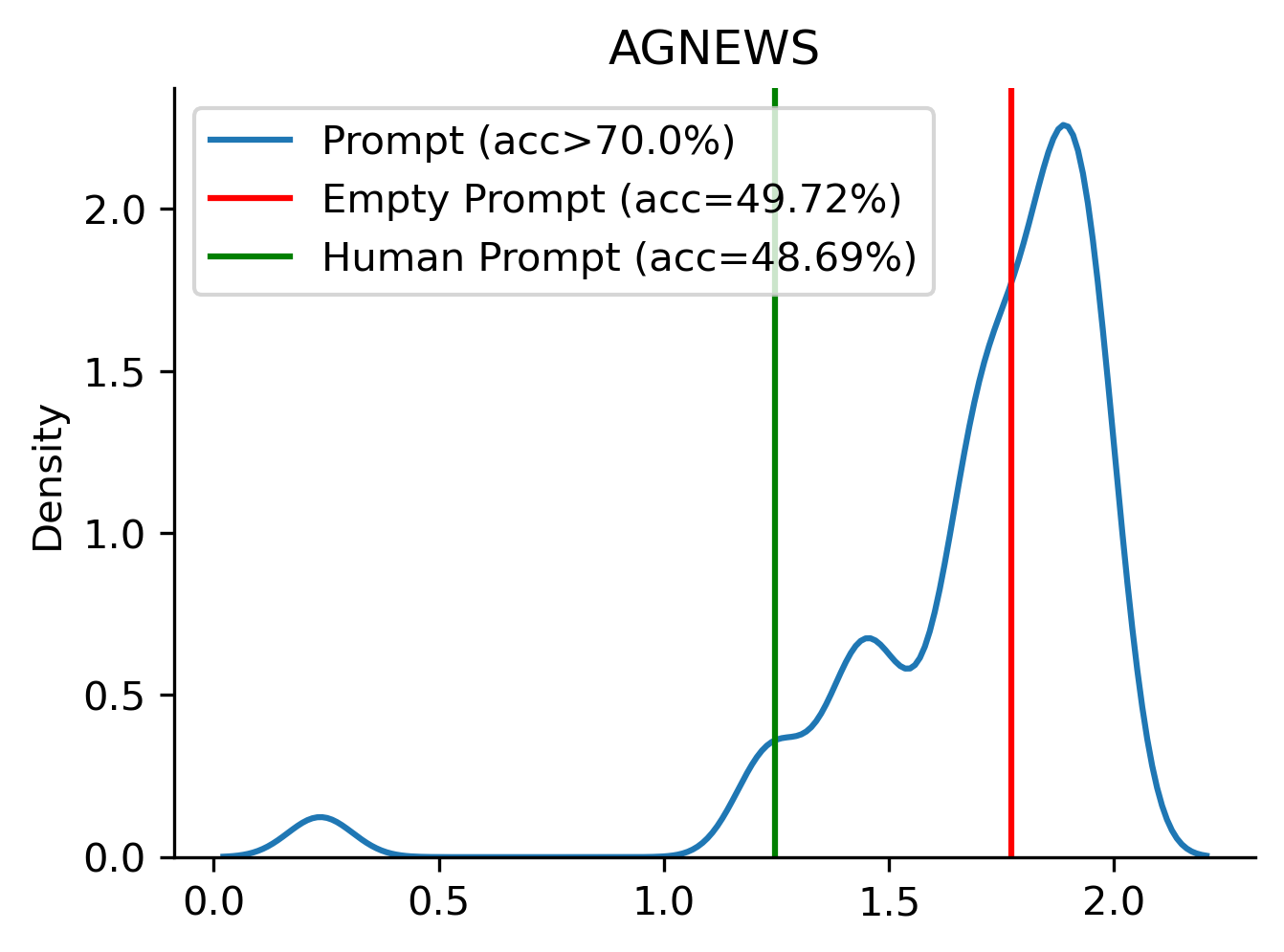}}
    \caption{
    % \weijia{entropy for x-axis. change color, frequency} 
    Frequency of prompts (y-axis) at different entropy level (x-axis). We compare effective prompts with the empty and human-written prompt. 
    % The effective prompt usually have high accuracy than empty and human-written prompt. For example, in SST-2, all performant prompts have accuracy greater than 80\%, while empty and human prompts have accuracy of 66.5\% and 55.9\% respectively.
    }
    \label{fig:density}
\end{figure*}

\begin{figure*}
    \centering
    \subfigure[]{\includegraphics[width=0.325\textwidth]{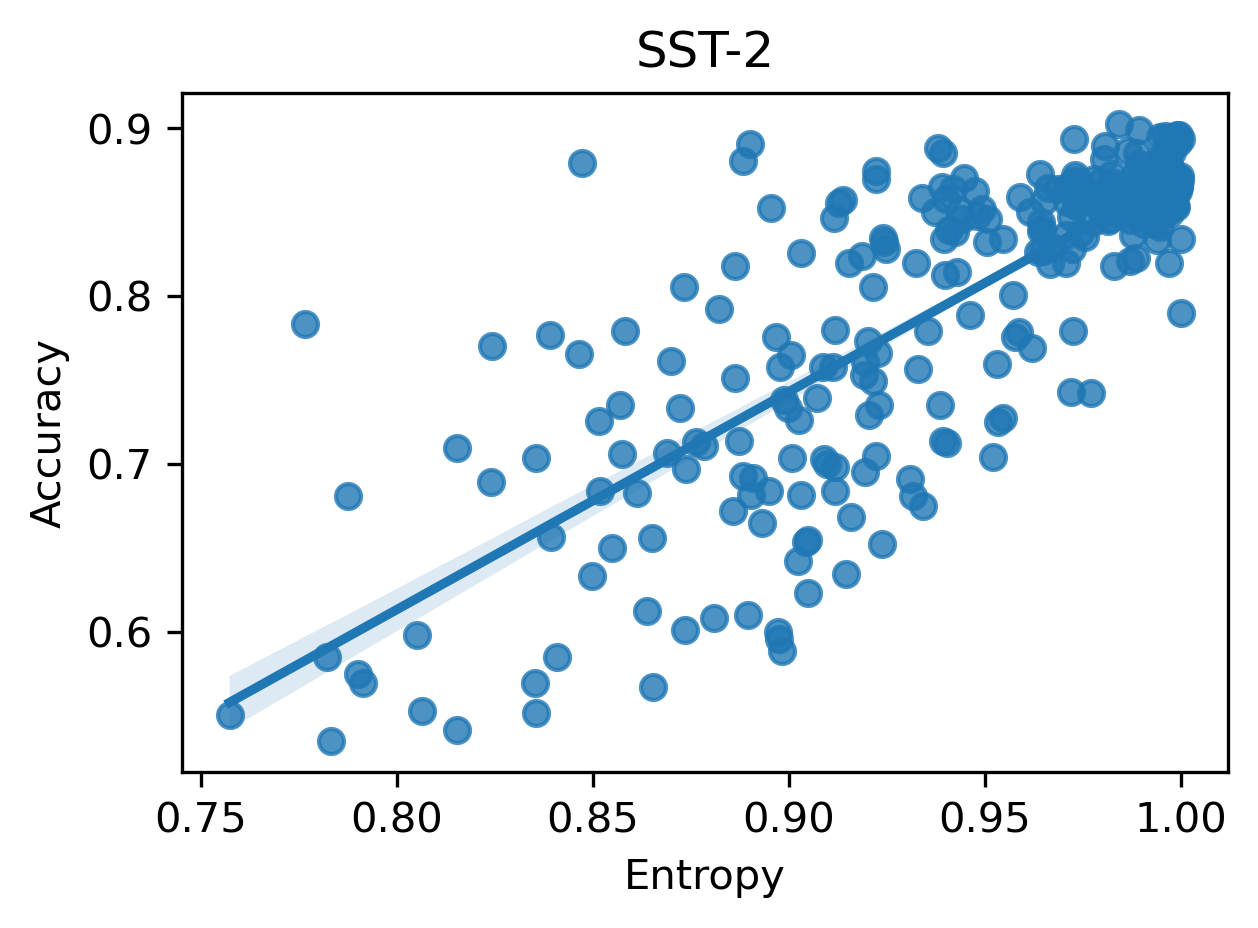}} 
    \subfigure[]{\includegraphics[width=0.325\textwidth]{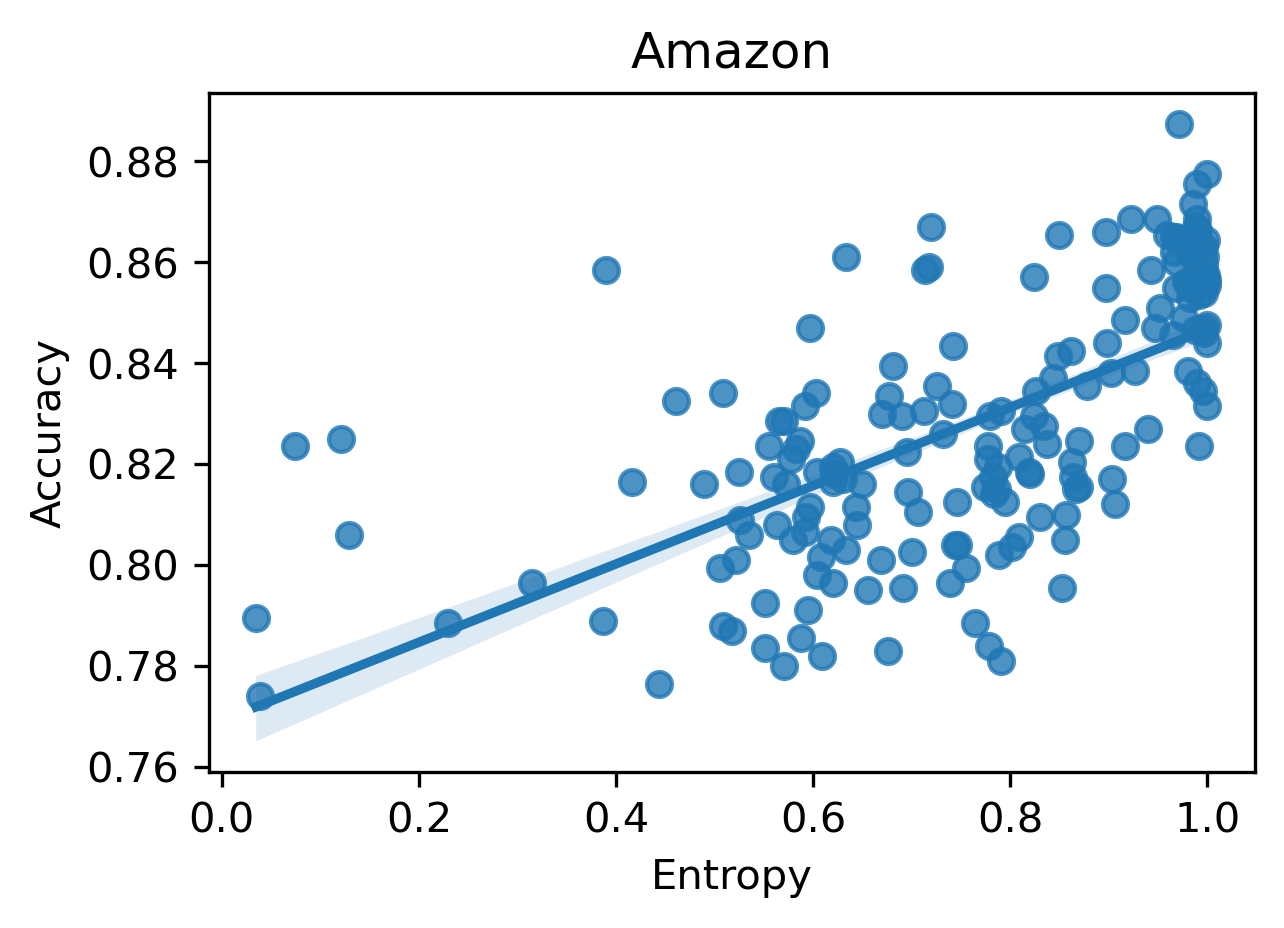}} 
    \subfigure[]{\includegraphics[width=0.325\textwidth]{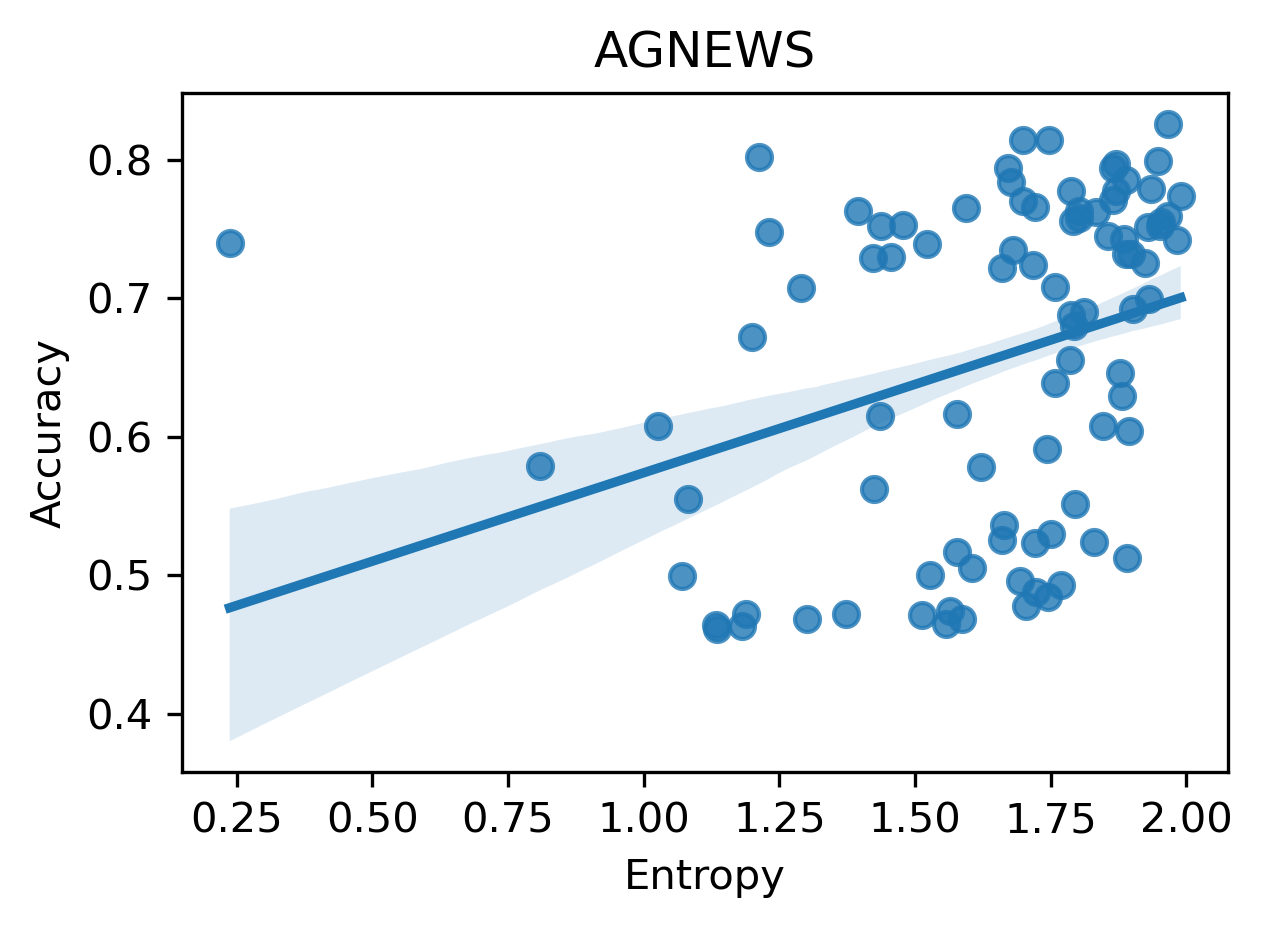}}
    \caption{Correlation between task performance and label word entropy. Spearman rank correlation coefficients for SST-2, Amazon and AGNEWS are +0.61, +0.75 and +0.43. All p-values are less than 0.0001.}
    \label{fig:entropy correlation}
\end{figure*}

% \subsection{Effective prompts calibrate the label word distribution}
\subsection{Effective prompts calibrate the output distribution over label words}
\label{sec:calibration_analysis}
% \hg{remove "word"?}
Language models are known to be biased towards label words that are common in its pretraining distribution~\cite{holtzman-etal-2021-surface, zhao2021calibrate}. 
% \han{In this section, we aim to investigate whether effective prompts found by prompt tuning implicitly adjust for the bias (calibration).}
In this section, we aim to investigate whether effective prompts found by prompt tuning implicitly adjust for the bias (calibration). 
To measure this bias, we follow \citet{holtzman-etal-2021-surface} to use task-specific domain string $\boldsymbol{d}$ as the test input and compute the entropy of the labels. As listed in Table \ref{tbl:domain string}, the task-specific domain strings do not imply any label information. Therefore, we expect the output of the language model to be more uniform over the label words when only conditioned on the domain string. The entropy of the label words is computed as follows:
\begin{table}
\small
\centering
\begin{tabular}{ll}
\toprule
Task & Domain String \\
\midrule
SST-2 & This is a movie review \\
Amazon & This is an Amazon product review \\
AGNEWS & This is a news \\
\bottomrule
\end{tabular}
\caption{Tasks and their task-specific domain strings.} \label{tbl:domain string}
\end{table}
\begin{align*}
&H(y) = \mathbb{E}_{y \in Y}[-\log p(y)] =\\
% -\sum_{y\in Y}{p(y) \log p(y)} \\
&-\sum_{y\in Y}{p_{\theta}(v(y) \mid \Tilde{\boldsymbol{e}}, \boldsymbol{d}, \boldsymbol{t}) \log p_{\theta}(v(y) \mid \Tilde{\boldsymbol{e}}, \boldsymbol{d}, \boldsymbol{t}) }
\end{align*}
The higher the entropy is, the more balanced (calibrated) the label words distribution is. 
% For example, in sentiment analysis task, if a language model is unbiased and scores the domain string as 50\% positive and 50\% negative, the entropy will be 1. However, GPT2-large outputs 75\% positive for "This is a movie review", substantially biased towards the positive label. 

As listed in Table \ref{tbl:prompt}, some effective prompts found by \methodname{} for sentiment analysis contain negative sentiment words (e.g., ``disappointing'' and ``complained'' in prompts for SST-2 ), which may implicitly reduce the probabilty of positive label and calibrate the label word distribution.  
To validate this hypothesis, we filter a set of effective prompts by \methodname{} and compute the entropy of the label predictions conditioned on the concatenation of the prompt and the task-specific domain string. Figure \ref{fig:density} shows 
% \hg{what is x axis? should be added to the plots} 
the density plot comparing the label word entropy of effective prompts, along with empty and human-written prompts taken from ~\citet{bach2022promptsource}. We observe that the entropy of effective prompts has a higher mode than the entropy of empty and human-written prompts with lower accuracy. 
% It suggests that performant prompts tend to calibrate the label word distribution. 

To further explore the relation between the task performance and calibration, we compute correlation between the task accuracy and the label word entropy of all prompts obtained by our algorithm \methodname{} and report Spearman's rank correlation. From Figure \ref{fig:entropy correlation}, we observe that the label word entropy exhibits significant positive correlations with the task accuracy (all $p<$ 0.0001). The Spearman's coefficients are +0.61, +0.75 and +0.43 for SST-2, Amazon and AGNEWS, respectively.

\subsection{Effective prompts are topically related to the task domain}
\label{sec:domain_analysis}

\paragraph{Qualitative Analysis}
% \han{have we mixed prompts with different lengths in Table 4? for SST-2 and Amazon it's short, for Agnews it's long?}

As shown in Table~\ref{tbl:prompt}, most of the effective prompts obtained by \methodname{} contain domain-related words. For example, the prompt \underline{\textit{Kubrick, "The Shining}} 
% and 
% \underline{\textit{Spielberg, The first time}} 
in SST-2 contains movie director names and movie titles, relevant to the domain of movie reviews. Similarly, the prompt \underline{\textit{mascara\textbackslash n\textbackslash n}} and \underline{\textit{cigars\textbackslash n\textbackslash n}} found for Amazon contain product names relevant to the domain of product reviews. Additionally, AGNEWS is a news topic classification task. Some of the effective prompts in AGNEWS contain topic classification-related words such as ``genre'', while others contain URLs that link to websites such as netflix\footnote{www.netflix.com} and pmwiki.\footnote{www.pmwiki.org} The target pages of these URLs also contain topic classification-related information, such as the prompt \underline{\textit{pmwiki/pmwiki.php/Main/Superhero}} which links to a \href{tvtropes.org/pmwiki/pmwiki.php/Main/Superhero}{wiki page} containing the following information: ``\emph{Genre: Action Adventure
Comedy
Commercials}''. 

\paragraph{Quantitative Analysis}
Based on our qualitative analysis, we hypothesize that effective prompts are topically related to the task domain. 
To validate this hypothesis, we compare domain word frequency in effective prompts and random sentences. 
First, we select a set of domain words for each task (see Table \ref{tbl:domain word}), which consist of the task label words (e.g., ``positive'' and ``negative'' for SST-2) and common words in the task domain (e.g., ``movie'' and ``film'' for the movie domain of SST-2). 
Since our prompts are very short (10 tokens), we augment each prompt with its continuation generated by GPT-3~\cite{brown2020language}, based on the assumption that the continuation by the large LM follows the same domain as the prompt. 
For each prompt, we sample 5 distinct continuations from GPT-3 using nucleus sampling $p=0.95$ at a length of 100 tokens. 
% Specifically, we sample 5 distinct continuations from GPT-3 for each prompt, with a length of 100 tokens. We use nucleus sampling~\cite{holtzman2019curious} with $p=0.95$ to generate continuations for the top 10 performance prompts. 
We compare the top 10 effective prompts with 10 random sentences from PILE~\cite{gao2020pile} augmented by the same continuations. 
% For comparison, we sample 10 random sentences from PILE~\cite{gao2020pile} to generate continuations. 
We then count the domain words in the concatenation of the prompt and its continuation. 

Table~\ref{tbl:domain frequency} lists the average accuracy of and number of domain words in the effective and random prompts with the continuations. 
% and the average  in the concatenation of the prompt and its continuation. 
The accuracy of effective prompts is higher than that of random sentences on all three datasets. Moreover, the domain words frequency of effective prompts is significantly higher than that of random sentences with p-values of 0.004, 0.003, and 0.0002 for SST-2, Amazon, and AGNEWS, respectively. 
Both our qualitative and quantitative analysis provide strong evidence that effective prompts obtained by our prompt tuning are topically related to the task's domain. 

% \han{mention drawbacks as well? we still don't have a systematic way to learn which specific domain-related words \emph{lead} to effective prompts}
% \weijia{not sure how to mention it. Maybe you can add something here?}

\begin{table}
\small
\centering
\begin{tabularx}{0.5\textwidth}{lX}
\toprule
Task & Domain Words \\
\midrule
SST-2 & movie, film, cinima, director, positive, negative \\
Amazon & book, amazon, product, furniture, positive, negative \\
AGNEWS & topic, category, politics, sports, business, technology \\
\bottomrule
\end{tabularx}
\caption{Domain words for each task. } \label{tbl:domain word}
\end{table}

\begin{table}
\small
\centering
\begin{tabular}{lllllll}
\toprule
& \multicolumn{2}{c}{SST-2} & \multicolumn{2}{c}{Amazon} & \multicolumn{2}{c}{AGNEWS} \\
\cmidrule(lr){2-3} \cmidrule(lr){4-5} \cmidrule(lr){6-7}
& Acc. & Freq. & Acc. & Freq. & Acc. & Freq. \\
\midrule
% Performant prompts & 89.4 & 23.35 $\pm$ 23.75 & 86.5 & 5.8 $\pm$ 4.3 & 77.6 & 3.7 $\pm$ 3.94 \\
% Random & 67.2 & 1.3 $\pm$ 1.41 & 74.2 & 2.2 $\pm$ 2.4 & 49.3 & 0.8 $\pm$ 1.16 \\

Effective & 89.4 & 23.4 & 86.5 & 5.8  & 77.6 & 3.7  \\
Random & 67.2 & 1.3  & 74.2 & 2.2 & 49.3 & 0.8  \\
\bottomrule 
\end{tabular}
\caption{Average domain words frequency (Freq.) and average accuracy (Acc.) for effective and random prompts. Effective prompts and their continuation contain subsantially more domain words than random prompts. 
The p-values from the paired t-test for SST-2, Amazon, and AGNEWS were 0.004, 0.003, and 0.0002, respectively.} 
\label{tbl:domain frequency}
\end{table}

% some of the performant prompts AG's news topic classification (AGNEWS) contain urls that link to netflix and pmwiki website. We carefully examine the target page of url and find that the page contains topic classification related information. For example, 

% Amazon: SpearmanrResult(correlation=0.6711606335066987, pvalue=3.5009800747149914e-25)

% SST2: SpearmanrResult(correlation=0.7564983604767445, pvalue=5.649137509569306e-51)

% Agnews: SpearmanrResult(correlation=0.4311430178240384, pvalue=1.9863636605719446e-05)

% \paragraph{Performant prompts with better readbility achieve better transferablity across models} 
% \weijia{test transfer between more models}
% \han{add dataset-transfer results in parallel to model-transfer results?}

\section{\newmethodname} \label{sec:new method}
% Unsupervised prompt tuning extending 
Our findings in Section \ref{sec:analysis} show the effective tuned prompts do calibration and have a high domain relevance to the task. These two attributes are both highly predictive and do not require ground-truth labels to compute. Therefore, in this section, we extend \methodname{} to explicitly tune the prompts towards better calibration and domain relevance, without using the task labels. 
% Following our domain relevance and calibration findings in the previous section, we extend \methodname{} to operate on \emph{unlabeled} training data of the task. We explicitly tune the prompts towards better calibration and higher domain relevance to the task/example. 
% \han{We assume we have a set of unlabeled training examples. We tune the prompts towards better calibration and domain relevance without using task labels.}

\subsection{Method}
\paragraph{Calibration loss}
In Section \ref{sec:calibration_analysis} we find a strong positive correlation between the degree of calibration (i.e., entropy) and performance of the prompts. We therefore explicitly optimize the prompt towards greater calibration, with an (negative) entropy loss defined below. 
\begin{align*}
\mathcal{L}_{\text{entropy}}(\Tilde{\boldsymbol{e}}) &= \mathbb{E}_{y \in Y}[\log \mathbb{E}_{\boldsymbol{x} \in X} p_{\theta}(v(y) \mid \Tilde{\boldsymbol{e}}, \boldsymbol{x}, \boldsymbol{t})]
\end{align*}
Intuitively the entropy loss encourages the prompt to help model generate more balanced predictions at a group level. 

\paragraph{Domain relevance loss}
In Section \ref{sec:domain_analysis} we find effective prompts overall are more related to the task domain, as defined by augmented data and keyword matches. To explicitly incorporate the domain relevance to the prompts, we extend the existing fluency (perplexity) loss in Section \ref{sec:fluency_constraint}, modeling the perplexity of both the prepending prompt and the input example:
\begin{align*}
    \mathcal{L}_{\text{domain}}(\Tilde{\boldsymbol{e}}) = &-\log p_{\theta}(\Tilde{\boldsymbol{e}}_{0:M})\\
    &-\sum_i \log p_{\theta}(x_i \mid \Tilde{\boldsymbol{e}}, \boldsymbol{x}_{<i})\\
    &-\sum_j \log p_{\theta}(t_j \mid \Tilde{\boldsymbol{e}}, \boldsymbol{x}, \boldsymbol{t}_{<j})
\end{align*}
Intuitively, $\log p_{\theta}(\boldsymbol{x} \mid \Tilde{\boldsymbol{e}}) - \log p_{\theta}(\boldsymbol{x})$ would measure the pointwise mutual information between the task data $\boldsymbol{x}$ and the tuned prompt $\boldsymbol{x}$, with the part $\log p_{\theta}(\boldsymbol{x})$ not involved in the prompt optimization. 

Overall, our unsupervised energy function $\mathcal{E}$ is updated to:
\begin{align*}
    \mathcal{E}(\Tilde{\boldsymbol{e}}_{0:M}) = &- \lambda_{\text{calibration}} \mathcal{L}_{\text{entropy}}(\Tilde{\boldsymbol{e}})\\
    &- \lambda_{\text{domain}} \mathcal{L}_{\text{domain}}(\Tilde{\boldsymbol{e}})
\end{align*}
where $\lambda_{\text{calibration}} + \lambda_{\text{domain}} = 1$.

\paragraph{Hyperparameters}
% We follow \methodname{} and use a learning rate $\eta \in$ \{1.0, 3.0\}, $\beta_{\text{start}} =$ 1.0, $\beta_{\text{end}} =$ 0.0001, $\lambda_{\text{domain}} \in$ of \{0, 0.0003, 0.001, 0.003, 0.01, 0.05, 0.2, 0.5\}, $M =$ 5 as the hyperparameters. We also use five random seeds for each setup. We select the best learning rate based on the validation set performance and report model performance averaged over the five random seeds.

Inheriting the notations of \methodname{}, we consider the following hyperparameters: $\eta \in$ \{1.0, 3.0\}, $\beta_{\text{start}} =$ 1.0, $\beta_{\text{end}} =$ 0.0001, $\lambda_{\text{domain}} \in$ \{0, 0.0003, 0.001, 0.003, 0.01, 0.05, 0.2, 0.5\}, $M=$ 10. We use five random seeds for each setup and report the average performance. 

% In our experiment, we follow \methodname{} method to perform a hyperparameter search to identify the optimal values for the learning rate ($\eta$), $\beta$, $\lambda_{\text{domain}}$, and $M$. We consider $\eta$ values of 1.0 and 3.0, set $\beta_{\text{start}}$ to 1.0 and $\beta_{\text{end}}$ to 0.0001, and test a range of values for $\lambda_{\text{domain}}$ including 0, 0.0003, 0.001, 0.003, 0.01, 0.05, 0.2, and 0.5. For each combination of hyperparameter setting, we use five random seeds.
% % We fix $M$ at 5 and use five random seeds for each combination of hyperparameters. 
% The best learning rate is selected based on the validation set performance, and the model performance is reported as the average across the five random seeds.

% \subsection{Imp}
\subsection{Results}
In Table~\ref{tbl:unsupervised prompt}, we compare the performance of our proposed method, \newmethodname{}, with two other unsupervised methods, the empty prompt and PMI calibration $\text{PMI}_\text{DC}$~\cite{holtzman-etal-2021-surface} on three datasets. 
% All of these methods do not require any labeled data.
Our results show that \newmethodname{} consistently outperforms $\text{PMI}_\text{DC}$ with an average improvement of 7.0\% across the datasets. This demonstrates the incorporated calibration and domain information 
are helpful to finding effective prompts. 
% are effective in creating natural language prompts. 
\begin{table}[t]
\centering
\small
\begin{tabular}{lcccc}
\toprule
& SST-2 & Amazon & AGNEWS\\
\midrule
\textbf{Unsupervised} \\
% Emtpy & 66.5 & 75.8 & 49.7 & 48.2 \\
% $\text{PMI}_\text{DC}$ & 85.6 & 76.2 & 64.1 & 50 \\
% \textsc{Unsup. R.P.}  & \textbf{88.3} & \textbf{85.5} & \textbf{68.5} & \textbf{60.7 }\\
Emtpy & 66.5 & 75.8 & 49.7 \\
$\text{PMI}_\text{DC}$ & 85.6 & 76.2 & 64.1 \\
\textsc{Unsup. F.P.}  & \textbf{88.2} & \textbf{85.3} & \textbf{68.0} \\

% \textsc{ReadPrompt}  &  &  &  & \\
\bottomrule
% \makecell[c]{\textsc{Unsupervised} \\ \textsc{ReadPrompt}} 
\end{tabular}
\caption{Accuracy of different unsupervised prompting methods on the three datasets. \textsc{Unsup. F.P.} refers to our \newmethodname{}.  
% \weijia{1. I use the abbreviation of \newmethodname{} because of the table does not fit single column. 2. what are other baselines 3. @Han I add CB here, not sure if it is necessary (feel free to remove it)}
} \label{tbl:unsupervised prompt}

\end{table}

% ('/private/home/swj0419/i-am-a-dog/openprompt-clone/swj_logging_dir/new2/amazon/gpt2-large/headprompt{5}_ppl_loss_lambda{0.001}_ppl_loss_example_lambda{0.001}_task_loss_lambda{0}_entropy_loss_lambda{0.999}_LR{3.0}_NSTART{1}_NEND{0.0001}_SEED{15}', (0.8555, 'Rating Pleasant humid disgusting\n')) 439948.6875

\section{Conclusion}
% In this paper, we study what factors make prompts effective.
% To faliciate this study, we design a analysis-friendly human readable prompt tuning method (\methodname{}) and apply it to GPT-2 large to generate effective and readable prompt. 
% Our analysis shows that effective prompts are topically
% related to the task’s domain and calibrate the prior probability of label words. 

% Although the prompts discovered by \methodname{} are effective and human-readable, they are still not semantically meaningful. For example, we didn't find any prompt related to the task definition. One potential reason is that GPT-2 large model is not instruction-tuned. 
% Future work may explore applying \methodname{} to instruction-tuned model and see if they can automatically find instruction-like prompts. 

In this paper, we investigate the factors that contribute to the effectiveness of prompts. To facilitate this study, we develop a human-readable prompt tuning method \methodname{} and apply it to the GPT-2 large model to generate effective and readable prompts. Our analysis reveals that effective prompts are topically related to the task domain and calibrate the prior probability of label words. 

Although the prompts generated by \methodname{} are effective and readable, they still carry limited semantic meanings. For instance, we did not find any prompts directly indicating the task definition or instructions. One potential reason is that the GPT-2 large model is not instruction-tuned. Future work can apply \methodname{} to an instruction-tuned model to see if instruction-like prompts can be discovered. 

% \input{section/ack.tex}

% % what makes good prompts, Kubrick's the shining makes the sentiment classifier

\bibliography{anthology,custom}
\bibliographystyle{acl_natbib}

% \appendix

% \section{Example Appendix}
% \label{sec:appendix}
\newpage
% \newpage
% \newpage
% \onecolumn

\appendix
\begin{appendix}
\section{Verbalizer and templates}
Table~\ref{tbl:verbalizer} shows an example input, template and the verbalizer used for each task. 
\begin{table*}[ht]
\small
\centering
\begin{tabularx}{1\textwidth}{lXl}
\toprule
Task & Templates & Verbalizers \\
\midrule
SST-2 & Illuminating if overly talky documentary. \textcolor{red}{It was} & positive, negative \\
Amazon & Terrible service. \textcolor{red}{It was} & positive, negative  \\
AGNEWS & Economic growth in Japan slows down as the country
experiences. \textcolor{red}{It is about} & politics, sports, business, technology \\
\bottomrule
\end{tabularx}
\caption{The \textcolor{red}{template}, example (colored black) and verbalizer used for each dataset.} \label{tbl:verbalizer}
\end{table*}

\end{appendix}

\end{document}